\documentclass[10pt,twocolumn,letterpaper]{article}

\usepackage[pagenumbers]{cvpr} % To force page numbers, e.g. for an arXiv version

\definecolor{cvprblue}{rgb}{0.21,0.49,0.74}
\usepackage[pagebackref,breaklinks,colorlinks,allcolors=cvprblue]{hyperref}

\def\paperID{15409} % *** Enter the Paper ID here
\def\confName{CVPR}
\def\confYear{2026}

\title{From Tables to Signals: Revealing Spectral Adaptivity in TabPFN}

\author{
Jianqiao Zheng\textsuperscript{*} \qquad
Cameron Gordon\textsuperscript{*} \qquad
Yiping Ji \qquad
Hemanth Saratchandran \qquad
Simon Lucey\\[2mm]
Australian Institute for Machine Learning\\
University of Adelaide\\
{\tt\small \{first.last\}@adelaide.edu.au}\\[1mm]
\textsuperscript{*}Equal contribution
}

\usepackage[utf8]{inputenc} % UTF-8 input
\usepackage[T1]{fontenc}    % proper font encoding for accented characters

\usepackage{amsthm}

\newtheorem{proposition}{Proposition}
\newtheorem{remark}{Remark}
\newtheorem{definition}{Definition}

 \usepackage{algorithm}
\usepackage{algpseudocode}

\begin{document}
\maketitle

\begin{abstract}

Task-agnostic tabular foundation models such as TabPFN have achieved impressive performance on tabular learning tasks, yet the origins of their inductive biases remain poorly understood. In this work, we study TabPFN through the lens of signal reconstruction and provide the first frequency-based analysis of its in-context learning behavior. We show that TabPFN possesses a broader effective frequency capacity than standard ReLU-MLPs, even without hyperparameter tuning. Moreover, unlike MLPs whose spectra evolve primarily over training epochs, we find that TabPFN's spectral capacity adapts directly to the number of samples provided in-context, a phenomenon we term \emph{Spectral Adaptivity}. We further demonstrate that positional encoding modulates TabPFN's frequency response, mirroring classical results in implicit neural representations. Finally, we show that these properties enable TabPFN to perform training-free and hyperparameter-free image denoising, illustrating its potential as a task-agnostic implicit model. Our analysis provides new insight into the structure and inductive biases of tabular foundation models and highlights their promise for broader signal reconstruction tasks.
\end{abstract}    
\section{Introduction}
\label{sec:intro}

%%%%%%%%%%%%%%%%%%%%%%%%FRONT FIGURE%%%%%%%%%%%%%%%%%%%%%%%%%%%%%%%%%

%%%%%%%%%%%%%%%%%%%%%%%%FRONT FIGURE%%%%%%%%%%%%%%%%%%%%%%%%%%%%%%%%%

\begin{figure}
    \centering
    \includegraphics[width=1\linewidth]{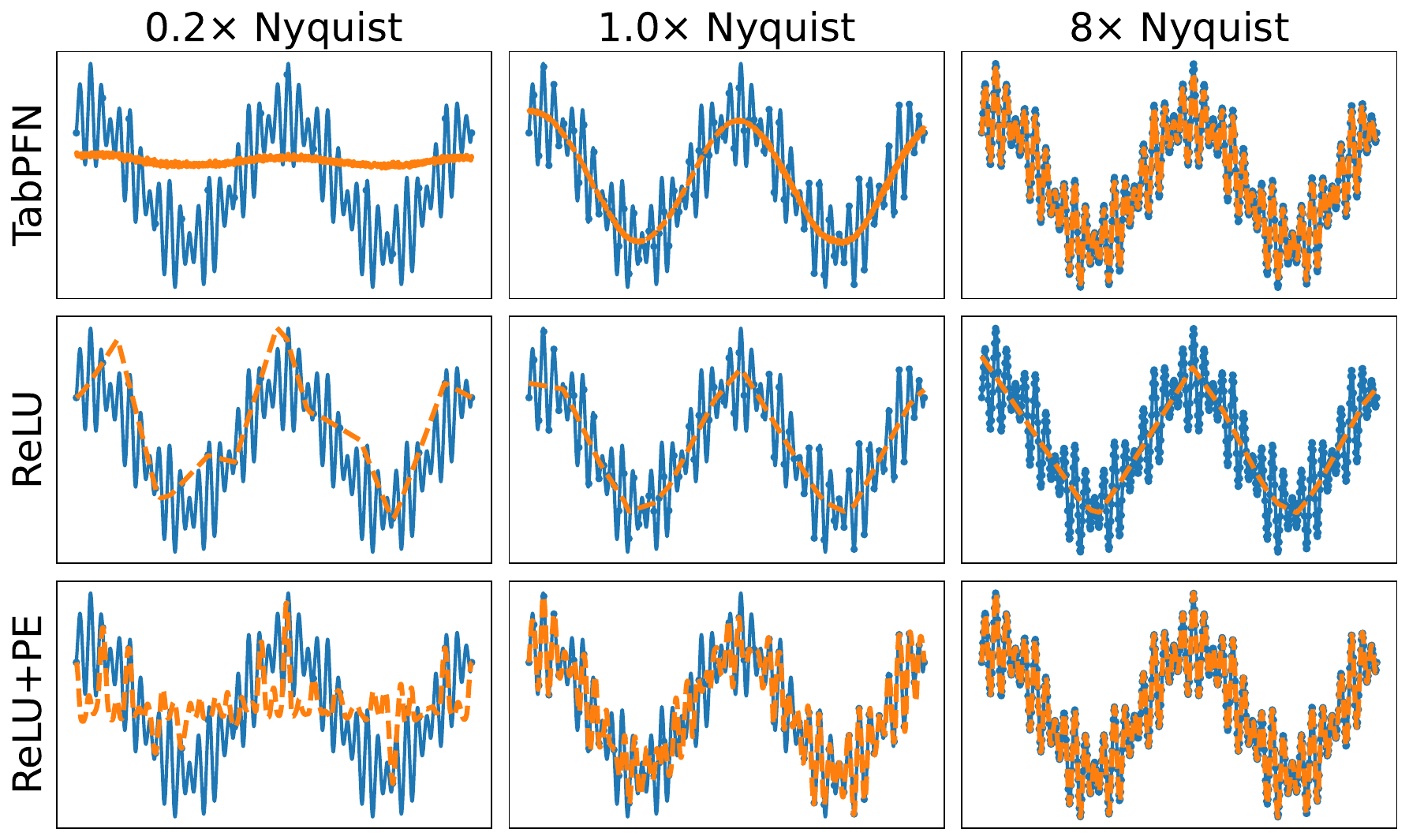}
    \caption{
    Signal reconstruction results at various sampling rates (columns) using TabPFN, ReLU-MLPs, and ReLU-MLPs with positional encoding (PE). ReLU-MLPs fail to reconstruct high-frequency components regardless of the amount of data provided, indicating that their spectrum is essentially insensitive to the number of samples. Positional encoding methods
    %(e.g., Random Fourier Features) 
    can extend their high-frequency capacity; however, hyperparameters tuned for one sampling rate (rightmost column) may cause unstable and noisy reconstructions at other sampling rates (leftmost column), requiring careful task-specific selection. In contrast, we find that the spectral capacity of TabPFN naturally adapts to the number of context samples—a phenomenon we refer to as \emph{Spectral Adaptivity}.
    }
\end{figure}

Task agnostic in-context learning approaches - of which TabPFN is a leading example - have recently revolutionized tabular machine learning \cite{jiang_representation_2025, jiang_representation_2025}. These Prior Fitted Networks (PFNs) trained on synthetic data are found to have strong performance without the need for selecting inductive biases \cite{hollmann2023tabpfn,hollmann2025tabpfn,muller2022transformers}. However, the properties of tabular foundation models are still not well understood - with most works currently treating them as a highly effective but mysterious black-box. In parallel, over the last decade implicit neural representations (INRs) have revolutionized the field of machine learning \cite{xie2022neural}. INRs seek to reconstruct signals implicitly through the trained weights of a neural network \cite{xie2022neural}. As they can be used to approximately reconstruct a fixed \textit{explicit} signal - such as a 1D signal or 2D image - they act as a useful test bed for understanding the theoretical inductive priors within neural networks, architectures, and optimizers \cite{saratchandran2023curvature, saratchandran2024activation}. By examining tabular foundation models through this lens we can demonstrate that in contrast to ReLU-MLPs whose spectral properties evolve during the course of gradient descent \cite{pmlr-v97-rahaman19a-spectral-bias}, TabPFN has a spectral bias that is controlled with the number of \textit{samples} provided to the in-context window. We show that - similar to work in implicit neural representations - that the inductive biases found in in-context models can be controlled with appropriate positional encoding. In this paper, we propose to provide a partial explanatory lens to TabPFN by examining these models through a signal reconstruction lens - bridging results with rapidly developing theoretical understanding in the properties of implicit neural representations.

Our paper has two audiences: the first is the tabular machine learning community, who may benefit from the bridge to theoretical tools for understanding the learning properties of tabular foundation models. Secondly, within the implicit neural representation community, it is often underappreciated that different signal domains necessitate models with specifically manipulated inductive biases to perform optimally. This involves an onerous, domain-specific search over architecture, hyperparameter, and optimizer configurations in order to achieve strong performance highlighting that no single `one-size-fits-all' model exists \cite{lucey2025gradientdescentshrinkageoperator}. For those we wish to highlight the potential for task-agnostic in-context models to be performant for signal reconstruction tasks without hyperparameter tuning. While the current context limitations of models render large implicit neural tasks - such as Neural Radiance Fields \cite{mildenhall2020nerf} - infeasible at the current time, we believe this path will be one of growing importance as research to overcome the limitations of context windows continues its rapid development \cite{sergazinov2025chunkedtabpfnexacttrainingfree, TabPFN-2.5}.

Our contributions include: 

% \begin{itemize}
%     \item[1.] We show that a signals processing lens can be applied to understand the performance of TabPFN's in-context learning. 
%     \item[2.] We show that - similar to MLPs - TabPFN exhibits a spectral bias for learning low-frequency signals. We show that positional encoding can be used to alleviate this bias. 
%     \item[3.] As a demonstration of this analysis, we show that TabPFN can be used to perform image signal processing tasks such as denoising. %, and classification. 
% \end{itemize}

\begin{itemize}
    \item[1.] We apply frequency-based signal analysis to examine the in-context learning behavior of TabPFN. Similar to ReLU-MLPs, we show that TabPFN exhibits a bias toward predicting low-frequency components, and that positional encoding can partially alleviate this bias.

    \item[2.] Unlike MLPs, we find that the spectral capacity of TabPFN naturally adapts to the number of context sample, a phenomenon we refer to as \emph{Spectral Adaptivity}.

    \item[3.] We show that TabPFN can be applied directly without fine-tuning to signal reconstruction tasks, enabling training-free and hyperparameter-free image denoising.
\end{itemize}

% \begin{itemize}
%     \item[1.] We perform frequency-based signal analysis to investigate the in-context learning behavior of TabPFN. Similar to ReLU-MLPs, TabPFN exhibits a bias toward predicting low-frequency components, which can be partially mitigated by positional encoding.

%     \item[2.] Unlike MLPs, the spectral capacity of TabPFN naturally adapts to the number of context samples. We refer to this phenomenon as \emph{Spectral Adaptivity}.

%     \item[3.] We demonstrate that TabPFN can be applied directly, without fine-tuning, to signal reconstruction tasks, enabling training-free and hyperparameter-free image denoising.
% \end{itemize}

% By exploring the inductive properties of TabPFN through a signal reconstruction lens we hope to provide a bridge between the rapidly emerging applied developments in tabular machine learning through the use of tabular foundation models, and the theoretical tools that have been used to understand the properties of implicit neural representations (INRs). Through this - we hope to provide some insights to techniques that have emerged from practical developments (such as the advantages of Fourier Features with TabPFN for time-series data \cite{hoo2025tablestimetabpfnv2outperforms}). 

\textbf{Paper overview}: Section 3 develops our theoretical framework, introducing the context kernel and formalising spectral adaptivity for in-context models. Section 4 then explores an expository 1D signal-reconstruction task as microscope to examine TabPFN’s inductive biases, contrasting it with ReLU-MLPs. Section 5 lifts this lens to a 2D inverse problem, building a training-free TabPFN pipeline for denoising and comparing it against standard INR baselines. Section 6 examines how positional encodings can act as frequency “dials” for TabPFN enabling it to learn higher frequency details with fewer samples - a result which provides insight to a recent methodological trend showing that positional encoding is required to use TabPFN for time-series data \cite{hoo2025tablestimetabpfnv2outperforms}), and Section 7 concludes with limitations and implications for future tabular and vision applications.

\section{Related Work}

\subsection{In-Context Tabular Learning} 

In-Context Learning (ICL) has been emerged as a key paradigm in which models perform tasks by analyzing prompts containing input-output pairs, without explicit training or parameter updates \cite{panwar2023incontext,brown_2020, dong2024surveyincontextlearning,xie2022explanationincontextlearningimplicit}. Recently, TabPFN - a family of in-context models that apply a transformer trained to approximate posterior predictive distributions over synthetic tasks - has emerged as the leading tabular foundational model for small datasets \cite{muller2022transformers, hollmann2023tabpfn, hollmann2025tabpfn, jiang_representation_2025, TabPFN-2.5, erickson2025tabarena}. These models have been applied to multiple tasks including time-series modeling ~\cite{hoo2025tablestimetabpfnv2outperforms}, reinforcement learning ~\cite{schiff2025gradientfreedeepreinforcement}, traffic forecasting ~\cite{li_architecture_2025}, and medical data analysis ~\cite{wang2024meditabscalingmedicaltabular}. Intriguingly, TabPFN has shown a remarkable ability to transfer without fine-tuning to new domains with properly encoded data. For example, ~\cite{eremeev2025turningtabularfoundationmodels, hayler2025of} find that TabPFN is able to solve graph problems formulated in a tabular format. In some cases, such as for time-series data, this has required the positional encoding or Fourier domain conversion of certain features for strong performance \cite{li_architecture_2025,hoo2025tablestimetabpfnv2outperforms, caiexplore}.  

The practical utility of TabPFN has meant that there is significant interest in understanding its properties beyond a black-box. However, to date there are only a papers exploring its properties. For example, by exploring the decision-boundaries in input spaces ~\cite{mccarter2025exactlytabpfnlearneddo}, the statistical properties of Prior-Fitted Networks ~\cite{pmlr-v202-nagler23a}, robustness to noise injection or corrupted data \cite{nawaz2025assessing,papastergios2150out}, or by examining how heterogeneity is handled \cite{ye2025a}. In addition, there is significant interest in scaling to larger context windows, either through architectural modification ~\cite{hollmann2025tabpfn,TabPFN-2.5,qu2025tabicl}, data reduction \cite{feuer2023scalingtabpfnsketchingfeature, ma2024incontextdatadistillationtabpfn, feuer2024tunetables}, or tiling and hierarchical inference strategies ~\cite{ye2025a,sergazinov2025chunkedtabpfnexacttrainingfree}.

\subsection{Signal Reconstruction}

Signal reconstruction is a core problem in vision and signal processing. The Shannon–Nyquist theorem \cite{shannon1949,Nyquist1928} states that a band-limited function with maximum frequency $\omega$ is fully determined by samples taken at rate $2\omega$. While this bound describes the worst-case sampling requirements for broadband signals, additional structure - such as sparsity or smoothness - can enable accurate recovery from substantially fewer samples \cite{brunton2022data}. INRs build on this principle of domain-agnostic approximation by parameterizing a continuous signal as a coordinate-based neural network \cite{xie2022neural, sitzmann2019siren,Occupancy_Networks,essakine2025standimplicitneuralrepresentations}. This framework has transformed several areas of computer vision: for instance, NeRF methods treat radiance fields as continuous functions and achieve high-quality novel-view synthesis from sparse observations \cite{mildenhall2020nerf, gao2025nerfneuralradiancefield}. In scientific machine learning, Physics-Informed Neural Networks (PINNs) extend this idea by embedding physical constraints directly into the coordinate-based model, enabling reliable reconstruction of solution fields even with extremely limited measurements \cite{RAISSI2019686,ramasinghe2023effectiveness}. A growing body of theoretical work seeks to explain why INRs are effective signal models - connecting activation functions, curvature constraints, sampling behavior, and invertibility to the types of frequencies a network can represent \cite{saratchandran2024samplingtheoryperspectiveactivations,saratchandran2023curvature,saratchandran2024activation,chng2024invertible}. These analyses provide a principled lens for understanding how neural architectures implicitly shape the reconstruction of continuous signals. We adopt this reconstruction viewpoint for TabPFN. By recasting signals as tabular datasets and analyzing TabPFN through a sampling-theoretic lens, we connect its in-context learning behavior to the rapidly developing theory behind INRs and reveal how its frequency response adapts to the number of context samples.

\subsection{Spectral Bias in Neural Networks}

Neural networks are known to exhibit a simplicity or spectral bias: they tend to fit low-complexity, low-frequency structure before high-frequency detail \cite{cao2020understandingspectralbiasdeep,pmlr-v119-basri20a, fridovichkeil2022spectralbiaspracticerole}. This has been demonstrated both empirically ~\cite{pmlr-v70-arpit17a-simplicity-bias,xu-2019-f-principle} and supported theoretically via Fourier analysis for both ReLU ~\cite{pmlr-v97-rahaman19a-spectral-bias} and sigmoid MLPs \cite{xu2018understandingtraininggeneralizationdeep}. Related work on coordinate-based INRs ~\cite{ramasinghe2022frequency} shows that increasing network bandwidth can suppress low frequencies unless explicit priors (e.g., activation choices) are provided. Additionally, positional encodings have been analyzed from a frequency-domain perspective to reveal how they shape learned spectral content \cite{zheng2021rethinkingpositionalencoding}. Despite this growing literature, the spectral behaviour of tabular foundation models remains largely unexplored. Our work helps to address this gap.

% \subsection{Training-Free Tabular Transfer Learning} 

% \subsection{Domain Transfer}

% It has been observed that linear probing outperforms fine-tuning for out-of-distribution samples \cite{kumar2022finetuning}. 

% \subsection{Signal Reconstruction}

% \subsection{Fine-Tuning tabular foundation models} 

\section{Spectral Properties of TabPFN}

\begin{figure}
    \centering
    \includegraphics[width=\linewidth]{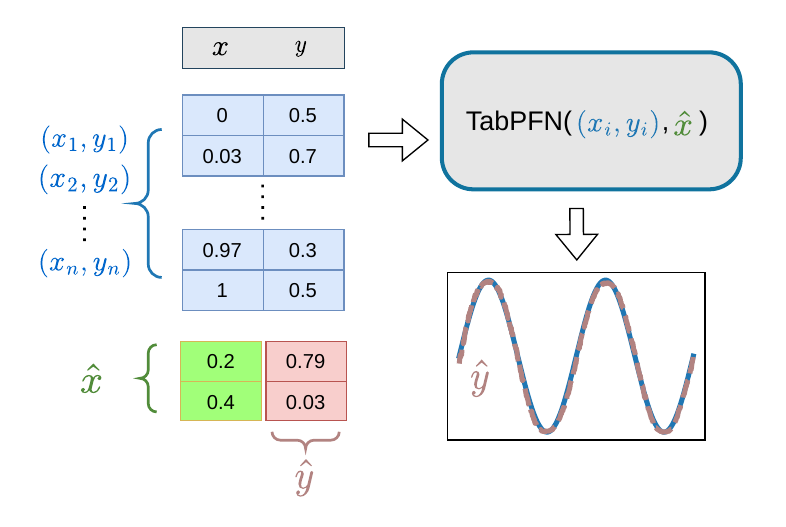}
    \caption{Recasting 1-D signal as tabular dataset. By representing sampled points $(x_i, y_i)$ as rows of a table, TabPFN can be applied to signal reconstruction: the model predicts $\hat{y}$ for query inputs $(\hat{x})$ in a single forward pass. Framing signals as tabular data enables analysis of TabPFN from a signal reconstruction perspective.}
    \label{fig:context_learning}
\end{figure}

In this section, we aim to theoretically compare the frequency response characteristics of optimization-based models (e.g., ReLU-MLPs) and in-context learning models (e.g., TabPFN). Building upon the concepts of Neural Tangent Kernel (NTK) and spectral bias for optimization-based models, we introduce the notions of \textit{context kernel} and \textit{spectral adaptivity} to describe in-context learning models.

\subsection{Problem Setup}

For simplicity, we consider a 1-D signal representation problem. Let the ground-truth function be $y = a(x),\, x \in [0, 1]$, and let the training dataset be $\mathcal{D}_{N} = \{(x_i, y_i)\}_{i=1}^{N}$. Our goal is to learn a model that approximates this function. 
Performance is evaluated using the mean squared error on a set of test (or query) points $q_j$.
\cref{fig:context_learning} illustrates how an in-context learning model addresses this problem.

\subsection{Preliminaries: Spectral Bias of ReLU-MLPs}

To address the 1-D signal representation problem using a ReLU-MLP model $f(x;\theta)$ trained via gradient descent, the standard objective is $\mathcal{L}(\theta) {=} \frac{1}{N}\sum_{i=1}^{N} \|f(x_i;\theta) - y_i\|^2$
Under the assumptions of the NTK regime~\cite{jacot2018neural}, the training dynamics can be approximated by kernel regression. 
The Neural Tangent Kernel (NTK) is defined as
\begin{equation}
    \label{equ:ntk}
    \mathbf{K}_{\mathrm{NTK}} = \big(\nabla_\theta f(\mathbf{X};\theta)\big)^{\top} \nabla_\theta f(\mathbf{X};\theta).
\end{equation}
Let the eigenvalue decomposition be $\mathbf{K} = \mathbf{Q}\mathbf{\Lambda}\mathbf{Q}^{\top}$. 
Then, the evolution of the function during training can be expressed as
\begin{equation}
    f_t(x) = \sum_{k} \big(1 - e^{-\lambda_k t}\big) \, b_k \, \phi_k(x),
\end{equation}
where $\lambda_{k}$ are the eigenvalues of the NTK, typically corresponding to the $k$-th frequency component $\phi_{k}(x)$, and $t$ denotes the training epoch.
During training, ReLU-MLPs tend to fit low-frequency components first and progressively capture higher-frequency details as optimization proceeds.

\begin{remark}[Spectral Bias~\cite{pmlr-v97-rahaman19a-spectral-bias}]
\label{remark:spectral_bias}
    The spectral bias of ReLU-MLPs arises from both the parameterization and the optimization process. 
    Given a fixed model architecture, initialization, and optimizer, the \textit{number of training epochs} effectively governs the evolution of this spectral bias.
\end{remark}

\subsection{Spectral Bias of ReLU-MLPs is Data Agnostic}

The NTK is an intrinsic property of a model that depends on its architecture (e.g., number of layers, width, skip connections, batch normalization) and initialization, as it is determined by the model’s gradients with respect to its parameters~\cite{jacot2018neural, golikov2022neuraltangentkernelsurvey}. 
We refer to these factors collectively as the \textit{parameterization} in~\cref{remark:spectral_bias}. 

While it is well known that the spectral bias of ReLU-MLPs evolves with training epochs, relatively few works~\cite{xiao2020disentangling} have emphasized that this spectral bias is largely independent of the number of training samples. 
Although the NTK is computed from the training samples, it serves only as a discrete approximation of an underlying continuous kernel. 
Increasing the number of samples merely refines this approximation; it does not alter the intrinsic spectrum of the NTK, which ultimately governs the spectral bias. 
% This implies that, regardless of how densely the training data samples the target function, the model cannot overcome its intrinsic limitation in representing higher frequencies.
This observation leads to the following proposition.

\begin{proposition}[Data‑Agnostic Spectral Bias in ReLU‑MLPs]
\label{proposition:data_irrelevance}
For a ReLU‑MLP with fixed architecture and initialization, the spectral distribution of its NTK, and consequently its spectral bias, remains invariant with respect to the number of training samples \(N\).
\end{proposition}

\begin{figure}
    \centering
    \includegraphics[width=\linewidth]{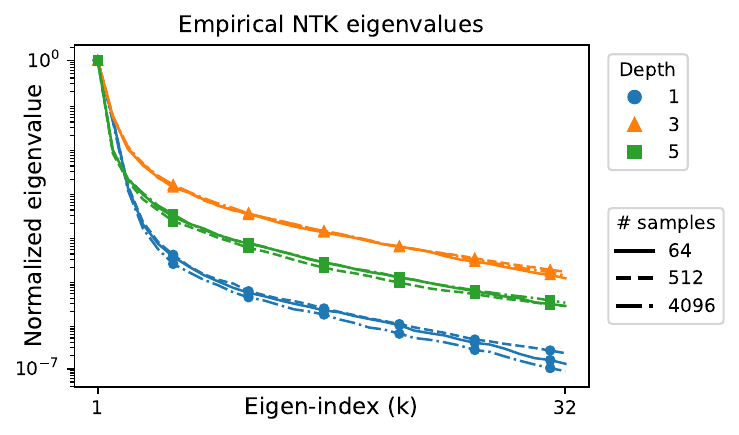}
    \caption{Normalized eigenvalue spectra of the empirical NTKs for ReLU-MLPs of varying depths, evaluated with 64, 512, and 4096 training samples. Across different sampling densities, the eigenvalue decay profiles remain unchanged, whereas network depth induces clear separations between curves. This demonstrates that NTK spectral properties are determined by the model architecture and are agnostic to the number of training samples.}
    \label{fig:ntk_eig}
\end{figure}

As an empirical validation, in~\cref{fig:ntk_eig}, we plot the eigenvalue spectra of NTKs for MLPs with 1, 3, and 5 layers, computed from $N {=} 64$, $512$, and $4096$ equally spaced samples in $[0,1]$. 
We observe that deeper MLPs exhibit distinct spectral profiles, while varying the number of training samples leaves the spectrum essentially unchanged.

\subsection{Preliminaries: In-Context Learning Model}
\label{sec:icl_prelim}

In-context learning (ICL) models, such as TabPFN~\cite{hollmann2023tabpfn}, typically employ a Transformer-based architecture. 
During pretraining, the model is exposed to a large number of synthetic tasks with known target functions, allowing it to learn a shared prior over the underlying data-generating processes. 
The pretraining objective is to predict the output for a new query \(x_*\) given a small context set \(\mathcal{C} = \{(x_i, y_i)\}_{i=1}^N\):
\[
\min_\theta \mathbb{E}_{p(\mathcal{C}, x_*, y_*)}\big[-\log p_\theta(y_* \mid x_*, \mathcal{C})\big].
\]
Once pretrained, the model parameters are fixed, and the model can be viewed as a function 
\(g(q;\mathcal{D})\) that performs in-context inference. 
Given a new context set $\mathcal{D}$, the model predicts outputs for query points $q$ via a single forward pass through the Transformer, without any gradient updates. 
% This enables generalization to unseen datasets and tasks purely through the model's learned in-context inference capability.

\begin{figure}
    \centering
    \includegraphics[width=\linewidth]{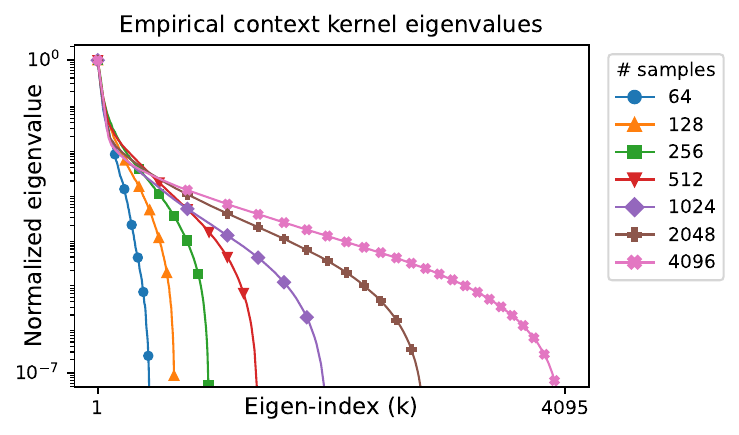}
    \caption{Eigenvalue decay profiles of the empirical context kernel for TabPFN across multiple sampling densities. The spectra clearly depend on the number of samples: higher sampling densities result in larger effective ranks and slower eigenvalue decay. This behavior contrasts with the sample-invariance observed in NTK spectra of classical MLPs, highlighting the sample-adaptive nature of the TabPFN context kernel.}
    \label{fig:ck_eig}
\end{figure}

\subsection{Context Kernel \& Spectral Adaptivity}
\label{sec:context_kernel}

In-context learning (ICL) models predict query outputs \(q_i\) through a single forward pass, without gradient-based parameter updates. 
Hence, computing the Neural Tangent Kernel (NTK), which relies on derivatives with respect to model parameters, is not meaningful for these models. 
However, the NTK conceptually measures how model predictions change in response to variations in training labels, mediated via network weights. 
Following this intuition, we define an analogous quantity for ICL models that directly captures the sensitivity of query outputs to context labels.

\begin{definition}[Context Kernel]
\label{def:context_kernel}
Given an in-context learning (ICL) model 
\(g: (\mathbf{X}_q, \mathbf{X}_c, \mathbf{y}_c) \mapsto \hat{\mathbf{y}}_q\), 
we define its \emph{context kernel} as the Jacobian of the predicted query outputs 
with respect to the context labels:
\begin{equation}
    \mathbf{K}_{\mathrm{context}} 
    := 
    \frac{\partial g(\mathbf{X}_q; \mathbf{X}_c, \mathbf{y}_c)}{\partial \mathbf{y}_c} \in \mathbb{R}^{M \times N},
\end{equation}
where \(M {=} |\mathbf{X}_q|\) is the number of query points and \(N {=} |\mathbf{X}_c|\) is the number of context points.
\end{definition}

The exact form of \(\mathbf{K}_{\mathrm{context}}\) is generally intractable. 
However, intuition can be gained from a single-layer attention block. 
Let the Transformer compute query outputs as
\begin{equation}
    \hat{\mathbf{y}}_q = \text{softmax}\Big( \frac{\mathbf{Q}_q \mathbf{K}_c^\top}{\sqrt{d}} \Big) y_c,
\end{equation}
where \(\mathbf{Q}_q {=} \mathbf{X}_q \mathbf{W}_Q\), \(\mathbf{K}_c {=} \mathbf{X}_c \mathbf{W}_K\).
The context kernel can be approximated as
\begin{equation}
    \label{equ:ck_attn}
    \mathbf{K}_{\mathrm{context}} = \frac{\partial \hat{\mathbf{y}}_q}{\partial \mathbf{y}_c} = \text{softmax}\Big( \frac{\mathbf{Q}_q \mathbf{K}_c^\top}{\sqrt{d}} \Big).
\end{equation}

Unlike ReLU-based MLPs, whose NTK remains fixed after initialization, 
the Transformer’s attention mechanism induces an input-dependent 
\(\mathbf{K}_{\mathrm{context}}\), which varies with the similarity structure and number of context samples \(N\).

\begin{figure*}[t]
    \centering
    \includegraphics[width=0.95\linewidth]{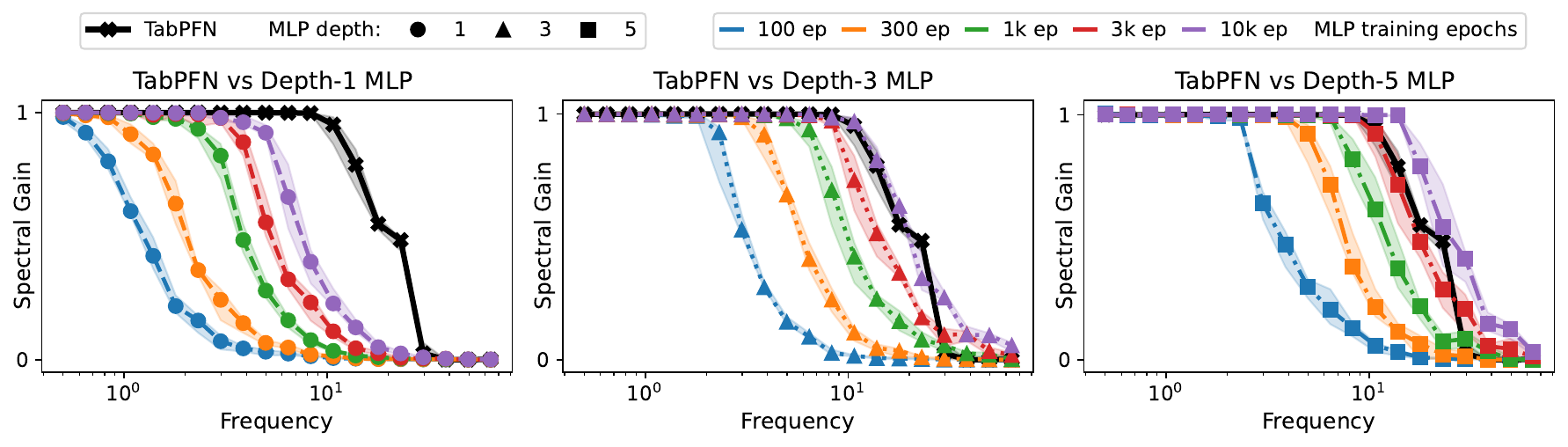}
    \caption{Spectral gain curves for TabPFN and ReLU MLPs with 1, 3, and 5 layers, evaluated across different training epochs. Each coloured curve corresponds to an MLP trained for a different number of epochs, illustrating how higher frequencies are learned progressively later during optimisation - a phenomenon commonly referred to as spectral bias. The black curve shows TabPFN’s spectral response, which already matches that of a well-trained deep MLP.}
    \label{fig:tab_mlp_epochs}
\end{figure*}

\begin{figure*}[t]
    \centering
    \includegraphics[width=0.95\linewidth]{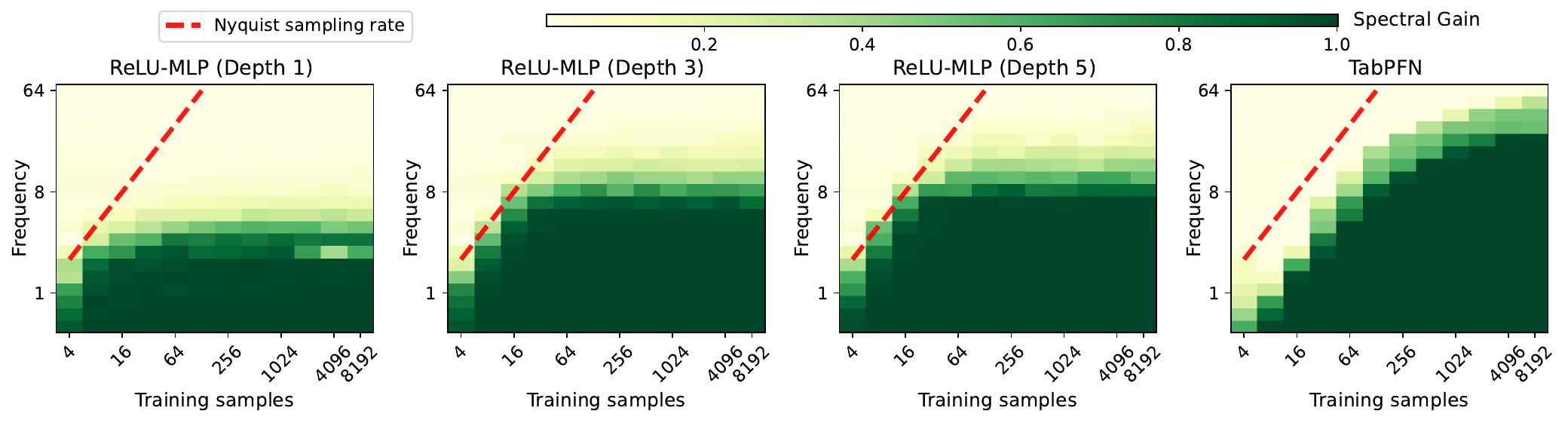}
    \caption{Spectral gain heatmaps for TabPFN and ReLU MLPs with 1, 3, and 5 layers, evaluated across different number of training samples. Darker regions indicate higher spectral gain (successful reconstruction). The red dashed line denotes the Nyquist sampling limit, above which no method can theoretically recover frequencies. The spectrum of MLPs does not change with the number of samples, except for low-frequency regions truncated by the Nyquist limit. In contrast, TabPFN continues to recover higher frequencies as the number of samples increases, demonstrating its spectral adaptivity.}
    \label{fig:tab_mlp_ntrain_heatmap}
\end{figure*}

\begin{proposition}
\label{prop:context_kernel}
The context kernel \(\mathbf{K}_{\mathrm{context}}\) of an ICL model, 
defined as the Jacobian of the predicted output with respect to the context labels, 
depends on the number of context samples \(N\). 
As \(N\) increases, its spectrum becomes progressively flatter and approaches the identity:
\begin{equation}
    \lim_{N\rightarrow\infty} 
    \mathbf{K}_{\mathrm{context}} 
    = 
    \frac{\partial g(\mathbf{X};\mathbf{X},\mathbf{y})}{\partial \mathbf{y}} 
    \rightarrow \mathbf{I}.
\end{equation}
\end{proposition}

To emperically validate this, in~\cref{fig:ck_eig}, we plot the eigenvalue spectra of context kernels for TabPFN computed from 
\(N = 64\), \(512\), and \(4096\) equally spaced samples in \([0,1]\). 
We observe that the spectrum flattens as \(N\) increases, in contrast to the near sample-invariance observed for NTKs of ReLU MLPs.

\begin{remark}[Spectral Adaptivity]
The frequency response of an in-context learning model (e.g., TabPFN) depends on both its parameterization and the number of context samples. 
Given a fixed pretrained model, increasing the number of context samples expands its range of representable frequencies. 
With few samples, the model captures only low-frequency components; with more samples, it can represent higher-frequency components. 
We refer to this phenomenon as \emph{spectral adaptivity}.
\end{remark}

\section{Synthetic Experiments}
\label{sec:synth_exp}

\begin{figure*}[t]
    \centering
    % tighten horizontal spacing if needed
    \setlength{\tabcolsep}{2pt}
    \renewcommand{\arraystretch}{0}

    \begin{tabular}{ccc}
        \includegraphics[width=0.32\textwidth]{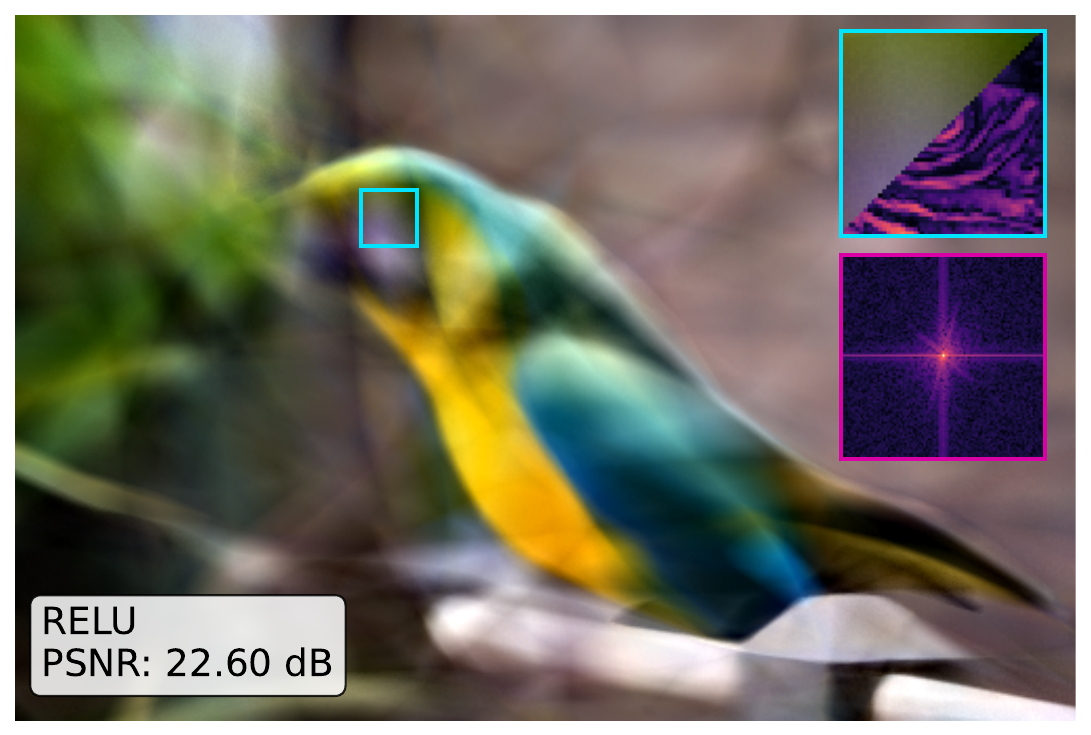} &
        \includegraphics[width=0.32\textwidth]
        {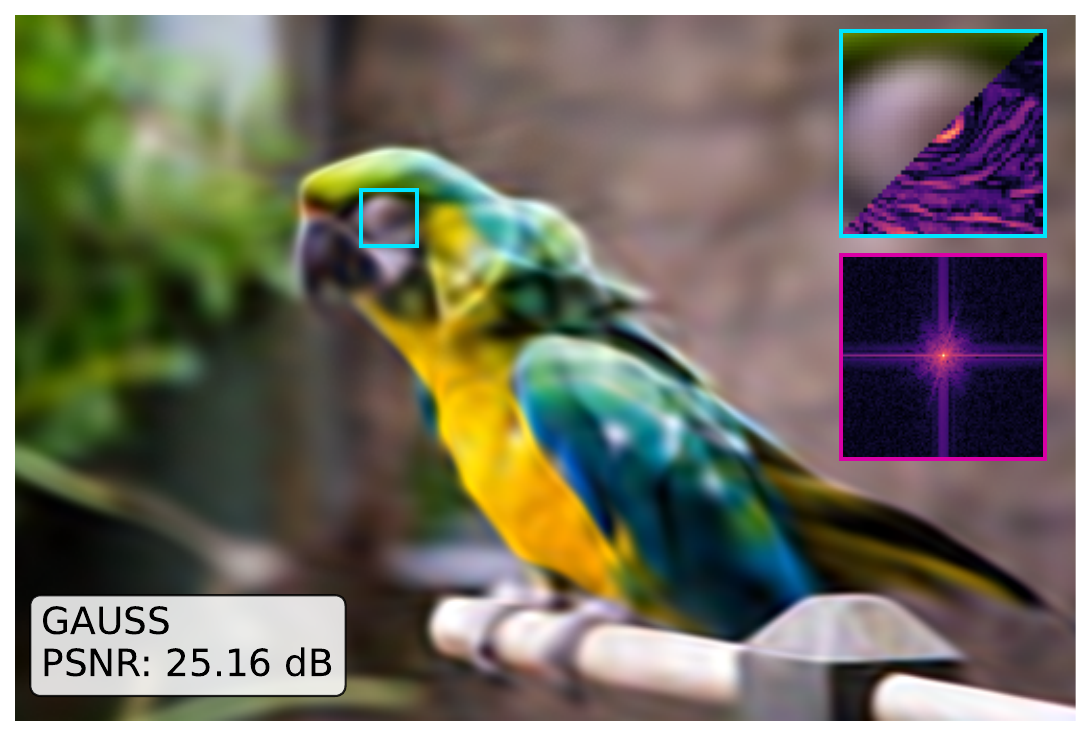} & 
        
        \includegraphics[width=0.32\textwidth]{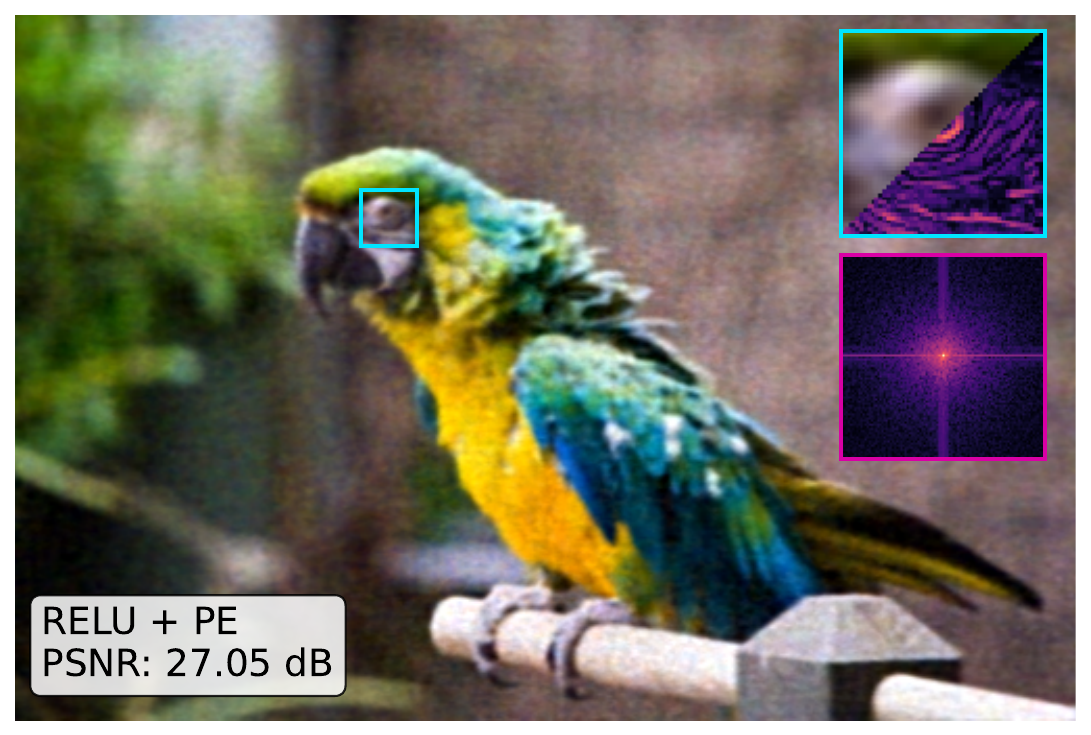} \\
\\

        (a) ReLU  & (b) Gaussian Activation \cite{ramasinghe2022} & (c)  ReLU+PE \\[4pt]

        \includegraphics[width=0.32\linewidth]{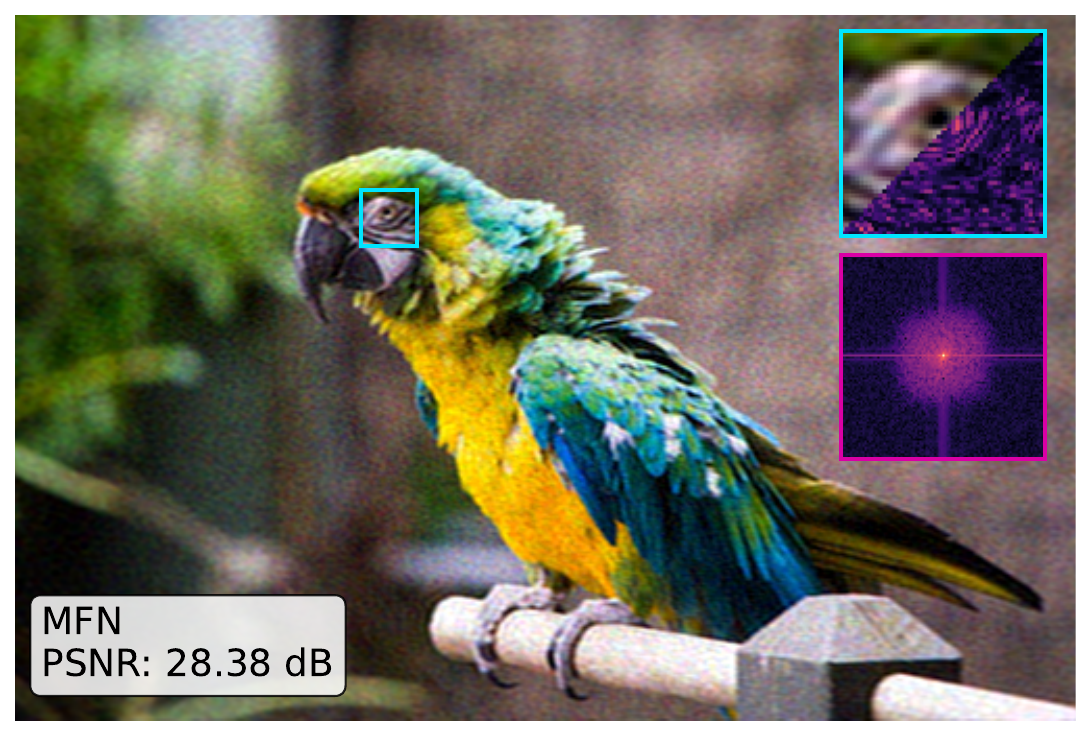} & 
                
        \includegraphics[width=0.32\textwidth]{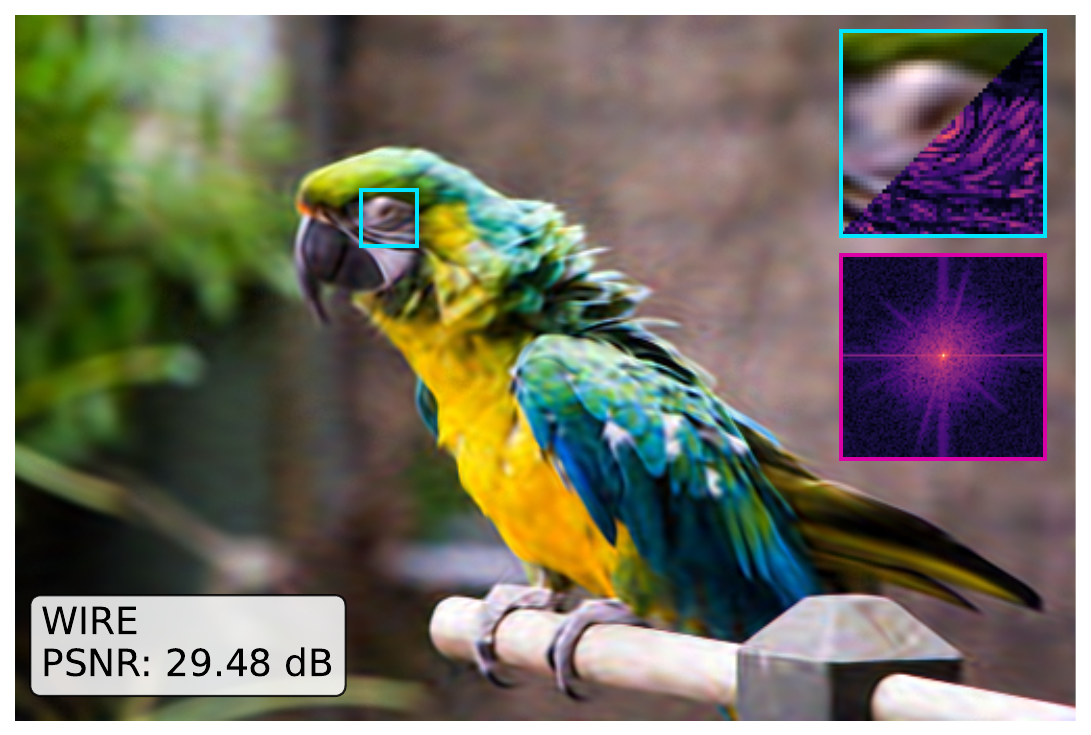} &
        \includegraphics[width=0.32\textwidth]{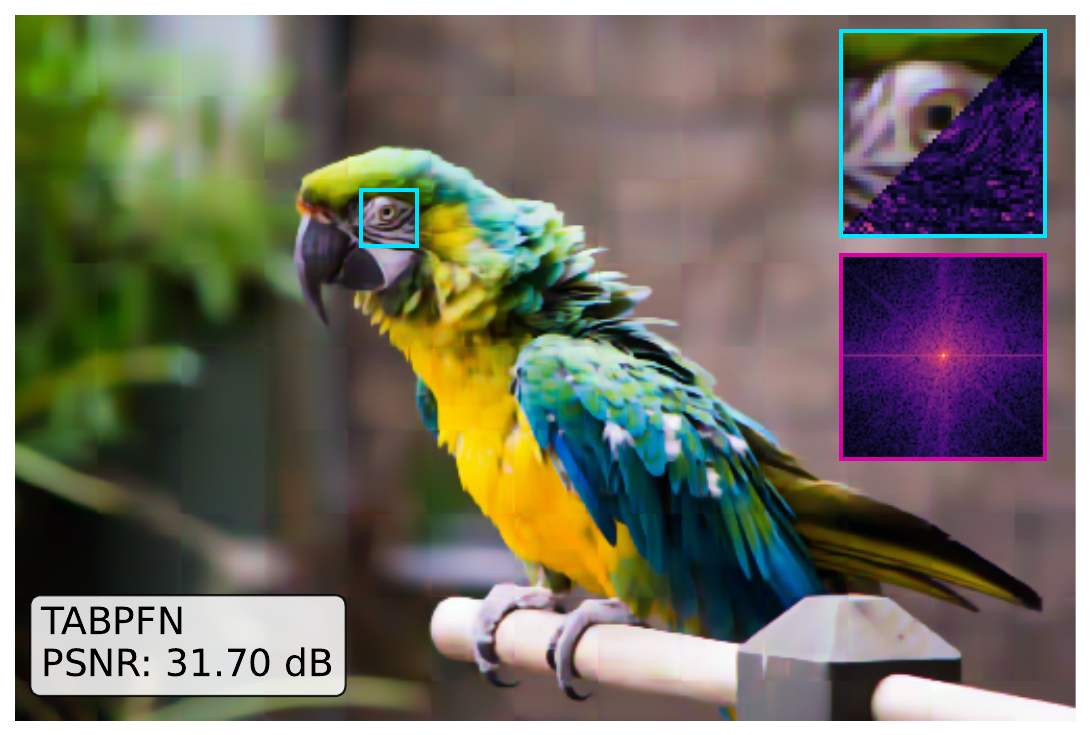} \\
        (d) MFN \cite{fathony2021multiplicative}  & (e) WIRE \cite{saragadam2023wire} & (f) TabPFNv2 \cite{hollmann2025tabpfn}
    \end{tabular}

    \caption{Comparison of neural implicit image reconstruction across common activation functions and architectures. Each subplot shows the reconstructed image of an example scene corrupted with noise. For each method, the top-right inset displays a spatial zoom highlighting local texture fidelity, and the bottom-right inset shows the log-magnitude 2D Fourier spectrum, revealing frequency content captured by the model. PSNR values (dB) are reported below each image. Overall, WIRE and TabPFNv2 recover sharper spatial details and preserve higher-frequency components, while ReLU variants exhibit pronounced low-frequency smoothing without positional encoding.}
    \label{fig:six-pdfs}
\end{figure*}

In this section, we use controlled synthetic experiments to analyse the spectral behaviour of TabPFN and ReLU MLPs, and to validate the proposed concept of \emph{spectral adaptivity}. Specifically, we study how their frequency responses evolve under varying numbers of training epochs (spectral bias) and sample densities (spectral adaptivity). We define the target function as
\begin{equation}
    y = \sin(2\pi kx + \varphi),
\end{equation}
where the frequency $k$ is geometrically sampled between $0.5$ and $20$, and the phase $\varphi$ is uniformly sampled in $[0, 2\pi]$. 
We compare TabPFN v2 against MLPs with $1$, $3$, and $5$ hidden layers of width $128$, trained using the Adam optimiser with a learning rate of $3\times10^{-3}$ and no weight decay. 
The output magnitude at each frequency component is reported as the \textit{spectral gain}.

\subsection{Varying Number of Epochs}

We first examine the effect of optimisation steps on the spectral behaviour of ReLU MLPs and TabPFN. 
Here, $256$ training samples are uniformly spaced between $0$ and $1$, and $2048$ test samples are used for evaluation. 
TabPFN does not involve gradient-based optimisation, so its result directly reflects its inherent spectral capacity, independent of training epochs. 
MLPs are trained for $100$, $300$, $1000$, $3000$, and $10000$ epochs. 
As shown in~\cref{fig:tab_mlp_epochs}, the frequency responses of MLPs gradually evolve with training, confirming their \emph{training-dependent} spectral bias. 
The spectral response of TabPFN is comparable to that of a well-trained deep MLP.

\subsection{Varying Number of Training Samples}

We next vary the number of training samples from $256$ to $8192$, uniformly sampled from $[0, 1]$, and evaluate on $16384$ test points. 
All MLPs are trained for $1000$ epochs. 
As shown in~\cref{fig:tab_mlp_ntrain_heatmap}, the spectral response of TabPFN continues to evolve as the number of training points increases, exhibiting stronger responses to higher frequencies as the sampling density increases. 
This stands in stark contrast to MLPs, whose spectral responses remain effectively unchanged. 
These results empirically validate our theoretical claim that in-context learning models possess \emph{spectral adaptivity}: their effective spectral bandwidth expands with the number of in-context samples.

\begin{figure*}
    \centering
    \includegraphics[width=0.95\linewidth]{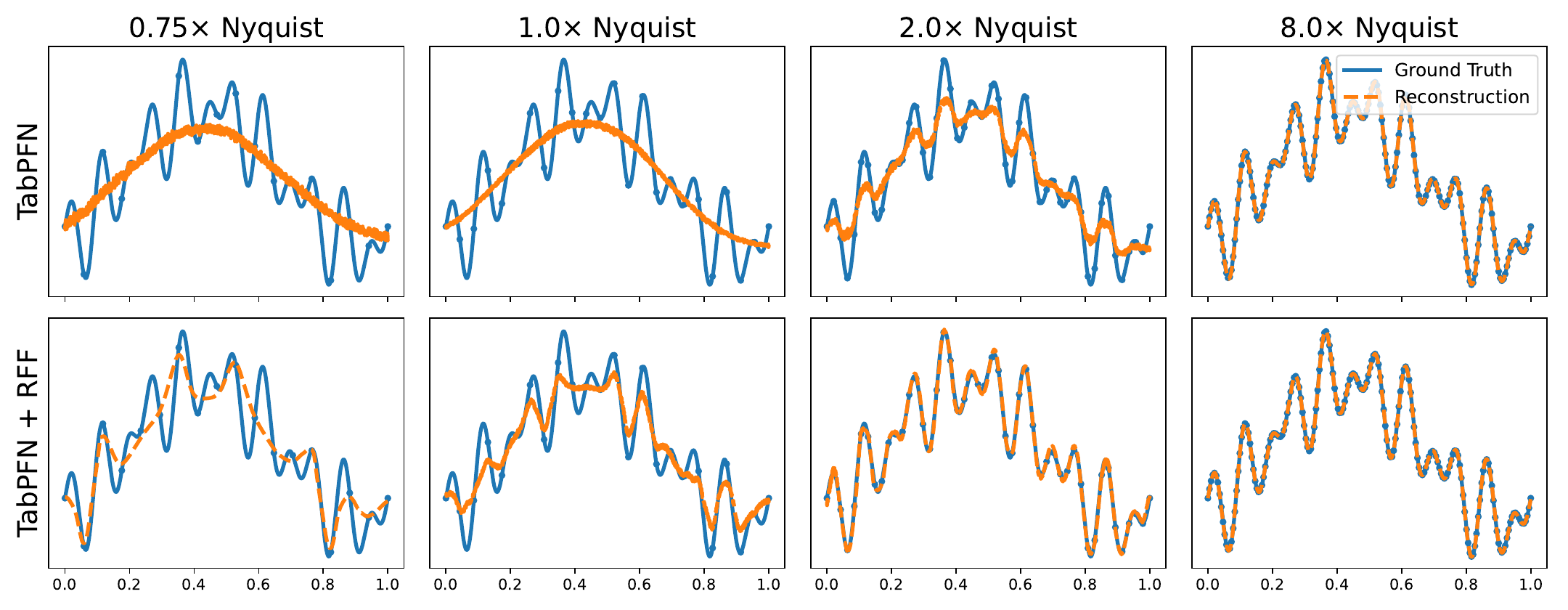}
    \caption{
        1D signal reconstruction across sampling rates. Top: TabPFN using raw coordinate inputs. Bottom: TabPFN with Random Fourier Feature positional encoding. As sampling density increases, TabPFN progressively recovers higher-frequency content. Fourier Features substantially improves reconstruction for sampling below the Shannon-Nyquist limit ($2f$ where $f$ is the maximum frequency).
    }
    \label{fig:nyquist}
\end{figure*}

\begin{figure*}[t]
    \centering
    \includegraphics[width=0.95\linewidth]{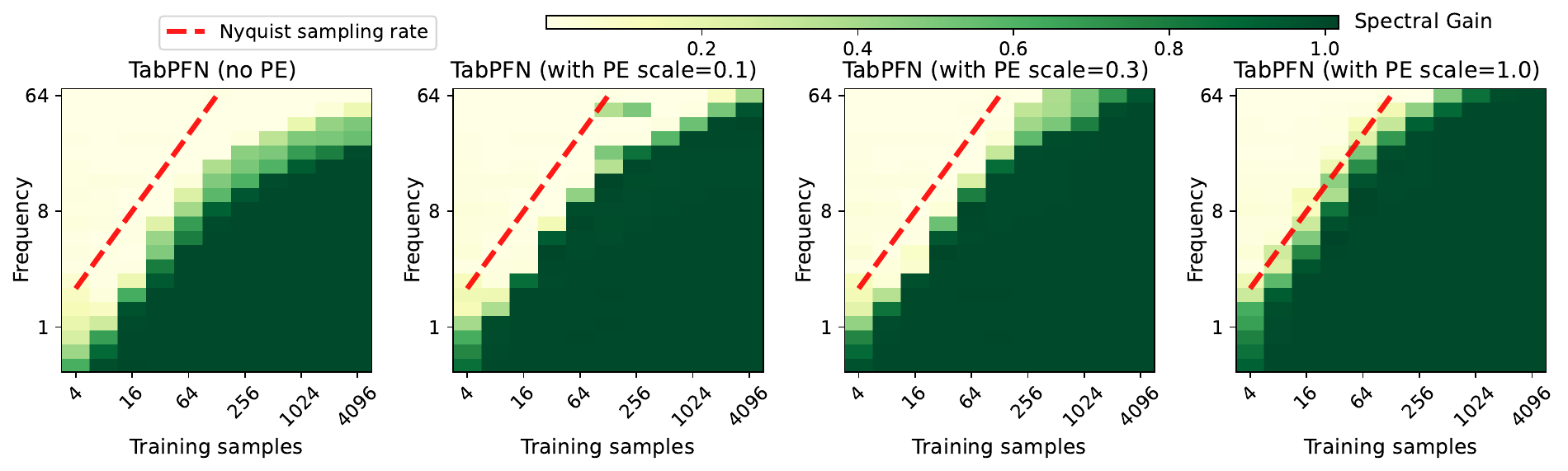}
    \caption{
        Effect of RFF positional encodings on TabPFN’s spectral behaviour. 
        Increasing the encoding scale enables TabPFN to recover progressively higher frequencies using fewer samples. 
        Low-scale encodings yield mild improvements over the no-PE baseline, whereas medium- and high-scale encodings substantially expand the recoverable bandwidth. 
        These results demonstrate that RFF encodings modulate TabPFN’s inductive bias and further enhance its spectral adaptivity.
    }
    \label{fig:tab_rff_heatmap}
\end{figure*}

\begin{figure}
    \centering
    \includegraphics[width=\linewidth]{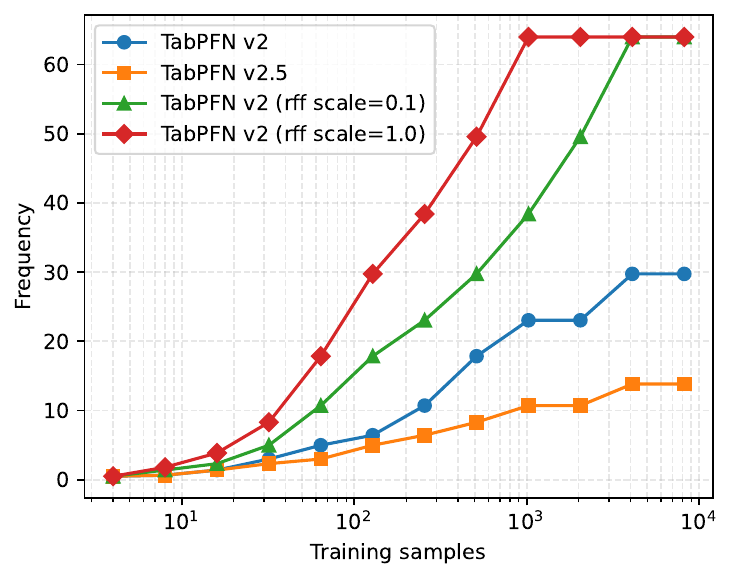}
    \caption{
        Frequency capacity of TabPFN under varying numbers of training samples. 
        We report the highest recoverable frequency (the first frequency with spectral gain $\leq 0.95$) for TabPFN v2, TabPFN v2.5, and TabPFN v2 augmented with RFF positional encodings of different bandwidths. 
        Spectral adaptivity, where recoverable frequency increases with sampling density, emerges as a general property of TabPFN-based in-context learning, and can be tuned through different pretraining strategies and input encodings.
    }
    \label{fig:v2.5}
\end{figure}

\section{TabPFN for Image Denoising}

\begin{algorithm}[t]
\caption{TabPFN Patch Denoising}
\label{alg:tabpfn-denoise}
\begin{algorithmic}[1]
\Require Noisy image $I$, patch size $P$
\Ensure Denoised image $\hat I$
\State $\hat I_\text{denoised} \gets 0$ 
\ForAll{patches $I_{\text{patch}}$} 
  % \State Coordinate matrix $X$ for the patch
  \For{each channel $ch \in {R, G, B}$}
      \State $y_\text{noisy} \gets$ pixel values of $I_{\text{patch, ch}}$
      \State Initialize TabPFN regressor $f$
      \State Fit $f$ on $(X_\text{coords},y_\text{noisy})$
      \State Predict denoised $\hat y = f(X_\text{coords})$
      % \State Write $\hat y$ back into channel $ch$ of $\hat I_{\text{patch}}$
      \State $\hat{I}_\text{denoised, patch, ch} \text{ = } \hat{y}$
  \EndFor
\EndFor
\State \Return $\hat{I}_\text{Denoised}$
\end{algorithmic}
\end{algorithm}

% \begin{figure}
%     \centering
%     % \includegraphics[width=0.32\linewidth]{Images/step_16_est_3_samp_1024_tot_262144.pdf}~
%     \includegraphics[width=0.45\linewidth]{Images/step_8_est_3_samp_4096_tot_262144.pdf}~
%     \includegraphics[width=0.45\linewidth]{Images/step_5_est_3_samp_10404_tot_262144.pdf}
%     \caption{Caption}
%     \label{fig:placeholder}
% \end{figure}

    % \paragraph{Image Denoising} 

    We adapt the image denoising pipeline from \cite{saragadam2023wire} to test TabPFN performance on a common signal reconstruction task. The denoising pipeline involves using a full set of training coordinates $x$ with labels corrupted by measurement noise $\hat{y}=y+\epsilon$. When trained under an implicit neural network the objective is to minimize the loss: 

    \begin{equation}
        \mathcal{L} = \|f_\theta(x)- \hat{y}\|
    \end{equation} 

    As perfect reconstruction including the additive noise is not desired, this acts as a test for the implicit regularization, inductive biases, and signal filtering behavior of different network types. We want to compare how TabPFN would perform under this scenario. We choose a (678, 1020, 3) target image from the WIRE test \cite{saragadam2023wire}. As the total number of pixels exceeds TabPFNv2's maximum context window, we employ a patching strategy described in \cref{alg:tabpfn-denoise} which and predict TabPFN separately for each set of coordinates and channel. Similar chunking strategies have recently been employed to extend TabPFN beyond its context window for tabular tasks \cite{sergazinov2025chunkedtabpfnexacttrainingfree}. For our purposes, this reflects an in-context forward pass for the tuple ($x_\text{patch}, \hat{y}; x_\text{patch}$) where the test and train data is the same. Note that no model training is occurring - the model remains fixed across the patches. In contrast, a single implicit neural representation would not be able to employ a similar patching strategy without costly retraining on each individual patch - or employing potentially dozens or hundreds of patch-specific models in a kiloNeRF style voxelization \cite{Reiser2021ICCV}. 
    % In order to reduce the potential for patching artifacts, we employ a soft overlap between each patch accumulating the overlapped predictions. 
    \cref{fig:six-pdfs} shows the qualitative results of the denoising. We can note that despite operating wholly through in-context predictions that TabPFN denoises favourably when compared to leading implicit neural representations including WIRE \cite{saragadam2023wire}, MFN \cite{fathony2021multiplicative}, Gaussian activated networks \cite{ramasinghe2022}, and RELU+Positional Encoding learned through gradient descent. In addition, we can note that TabPFN does not simply act as a naive low-pass filter - while it shows a spectral bias towards low-pass frequencies it still learns high frequency components, through its sampling consistent with our theoretical discussion.

\section{TabPFN with Positional Encoding}

Positional encodings (e.g., Random Fourier Features \cite{Rahimi_2007_random_fourier_features}) are widely used in implicit neural representations (INRs) to enrich the spectral expressivity of ReLU-MLPs \cite{ramasinghe2022frequency, tancik2020fourierfeaturesletnetworks}. 
Here, we investigate how positional encodings influence the spectral behavior of TabPFN. We begin with a 1D curve-fitting experiment using several types of positional encodings. 
We evaluate models under sampling densities both below and above the Shannon-Nyquist rate. 
As shown in \cref{fig:nyquist}, positional encodings substantially improve the approximation quality when the sampling rate is below Nyquist.
Next, we apply Random Fourier Feature (RFF) encodings with varying bandwidths, which modulate the effective input frequency content. 
As illustrated in \cref{fig:tab_rff_heatmap}, increasing the encoding scale systematically shifts TabPFN’s frequency response: larger bandwidths amplify high-frequency components and reduce the number of samples required to recover fine-scale details.
This demonstrates that TabPFN’s spectral adaptivity can be further adjusted through input-encoding design, analogous to frequency modulation techniques used in coordinate-based networks. Finally, \cref{fig:v2.5} plots the recoverable frequency versus the number of training samples. We additionally use this experiment to test the recently released TabPFNv2.5 \cite{TabPFN-2.5}
We observe that:  
(a) different pretrained TabPFN models (v2 and v2.5) exhibit distinct levels of spectral adaptivity, but both show the sample-based adaption consistent with our analysis;  
(b) different input frequency encoding scales can induce different spectral profiles; and  
(c) for a fixed parametrization, the recoverable frequency consistently increases with the number of samples. 
These results collectively highlight spectral adaptivity as a common attribute of TabPFN-based in-context learning.

\section{Conclusion and Limitations}

This work reframes TabPFN as a model whose inductive bias is not fixed by its architecture, but is instead shaped by the data provided in-context. Through the lens of the context kernel, we show that TabPFN exhibits \emph{spectral adaptivity} - its effective bandwidth expands as the sampling density of the prompt increases. This is in contrast to the architecture-governed spectral behavior of ReLU MLPs and INRs, whose frequency profiles remain largely unchanged once the model is specified. Across 1D reconstruction and image denoising, we find that TabPFN can recover increasingly high-frequency structure without iterative optimization, and that positional encodings serve as predictable mechanisms for steering its frequency response.

Our analysis inherits several limitations of the underlying model. TabPFN and related PFNs remain constrained by context-window size and input dimensionality ~\cite{TabPFN-2.5, qu2025tabicl}, which limits the spatial resolution and signal complexity of the vision tasks we can evaluate. The patching strategy used in our 2D experiments, while effective, introduces blending artefacts that are absent in globally trained INRs. Furthermore, our context-kernel interpretation relies on a local linearization of a nonlinear transformer, leaving open how spectral adaptivity emerges from full attention dynamics and the structure of the pretraining distribution.

Even so, the adaptive-spectral perspective suggests a broader design space for in-context models. If spectral behavior can be influenced by context geometry, positional encodings, and synthetic pretraining tasks - not solely by model architecture - then PFNs offer a path toward training-free signal models whose inductive biases evolve with the data they are given. Extending this idea to larger contexts, multi-scale visual signals, and temporally evolving prompts may connect tabular foundation models with implicit neural representations in a way that enables fast, adaptive reconstruction without gradient descent.

\section*{Acknowledgments}
This research publication was supported in part by the
CommBank Centre for Foundational AI Research.

% \subsection{Limitations} 
% There are two architectural limitations that constrain the use of in-context tabular foundation models in practical scenarios. In particular, they are constrained by the number of features and data samples that may be taken as context. For TabPFN this is currently around 500 features, and 10,000 data samples. TabICL can take more data samples (approximately 100,000), but is still relatively constrained on feature inputs. The method we have proposed has sought to practically reduce this limitation through the use of pre-trained encoders, however we note that this limitation still provides a significant constraint relative to the state-of-the-art in vision modeling.  

% \subsection{Future Work} 

% \clearpage

{
    \small
    \bibliographystyle{ieeenat_fullname}
    \bibliography{main}
}

\clearpage

\appendix
\setcounter{page}{1}
\setcounter{figure}{0}
\maketitlesupplementary

\section{TabPFN Attention Structure}
\label{app:attention_analysis}
A key question left open by the context-kernel analysis is how TabPFN’s internal representations change as more in-context samples are provided. In~\cref{sec:context_kernel}, we defined the context kernel and plotted its empirical version for TabPFN in Main Paper~\cref{fig:ck_eig}, since the true kernel is generally intractable. As TabPFN is Transformer-based, its context kernel depends on the self-attention mechanism shown in~\cref{equ:ck_attn}. We therefore examine the model’s attention patterns directly, as they are the main mechanism through which TabPFN modulates interactions between context points and thus its spectral capacity. This section provides results examining how TabPFN’s attention maps ($n \times n$) evolve with the number of in-context samples. 
These visualizations complement the main paper by showing how internal attention mechanisms become sharper and more expressive as the sampling density increases.

\subsection{Spectral Structure of Attention Maps via SVD}
\label{app:attn_svd}
\begin{figure}[t]
    \centering
    \includegraphics[width=0.95\linewidth]{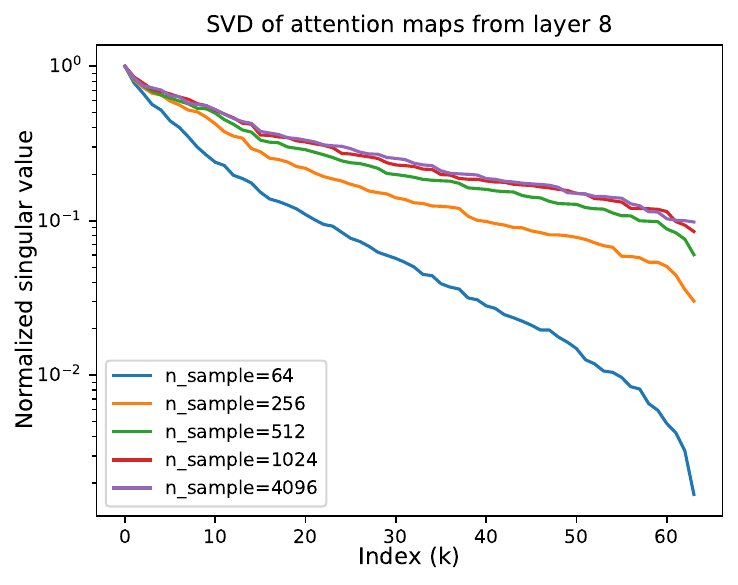}
    \caption{
        Singular value decomposition of Layer~8 self-attention maps (subsampled to same size of $64\times64$ from sample size) computed over different numbers of context samples ($64$–$4096$). 
        As shown in \cref{fig:tab_attn_map}, the attention maps become increasingly sharp and structured with more samples. 
        The SVD results provide a quantitative view of this trend: attention maps obtain progressively richer spectra as the number of context samples increases, indicating higher-rank and more expressive attention patterns.
    }
    \label{fig:attn_map_svd}
\end{figure}

To obtain a quantitative view of how TabPFN’s internal interactions become more expressive as the context grows, we analyse the spectral structure of its attention maps via singular value decomposition (SVD). \cref{fig:attn_map_svd} reports the singular value spectra of Layer~8 attention maps across the same range of sample sizes. The singular value spectrum broadens with additional in-context samples, indicating higher-rank attention maps and richer internal representations. This trend mirrors the rise in recoverable frequencies reported in the main text and further supports the interpretation of TabPFN exhibiting \emph{spectral adaptivity} at both representation and attention levels.

\subsection{Self-Attention Maps Across Layers}
\label{app:attn_maps}

\begin{figure*}[t]
    \centering
    \includegraphics[width=0.95\linewidth]{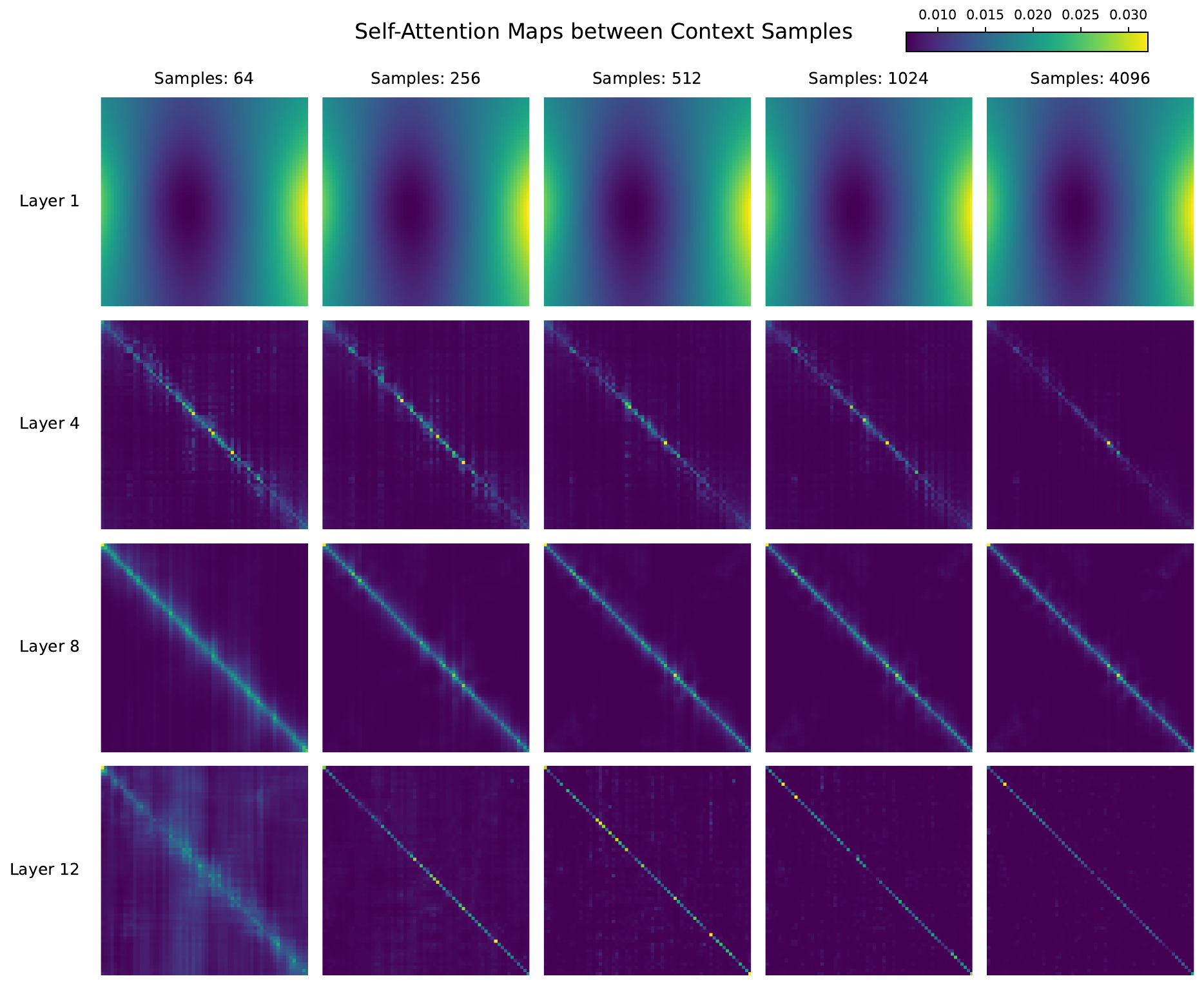}
    \caption{
        Self-attention maps of TabPFN across transformer layers and different numbers of context samples. 
        Columns correspond to context sample sizes ($64$, $256$, $512$, $1024$, $4096$), and rows correspond to Layers~1, 4, 8, and~12. 
        All attention maps are averaged over 6 heads and subsampled to $64\times64$ for consistent visual comparison. 
        As sampling increases, the attention maps become sharper, more structured, and more concentrated around local neighborhoods. 
        These qualitative changes reflect the model’s increasingly expressive internal representations and complement the quantitative SVD analysis in \cref{fig:attn_map_svd}, evidencing TabPFN’s sample-dependent attention behavior.
    }
    \label{fig:tab_attn_map}
\end{figure*}

We can visualize self-attention maps to observe whether increasing sample density leads to progressively sharper and more localized attention neighborhoods, as would be expected from a model whose inductive bias adapts to sampling resolution. \cref{fig:tab_attn_map} presents self-attention maps for layers~1, 4, 8, and~12 of TabPFN under context sample sizes ranging from $64$ to $4096$. All attention matrices are averaged over 6 heads and subsampled to $64\times64$ for consistent visual comparison. We observe that as the number of samples increases, the maps exhibit increasingly structured and localized patterns, reflecting the formation of more selective attention neighborhoods and aligning with the sample-dependent behavior observed in the spectral experiments.

% \begin{figure}[t]
%     \centering
%     \includegraphics[width=0.95\linewidth]{Images/results_time_comparison_1.pdf}
%     \caption{
%     }
%     \label{fig:patch_size_ablation_1}
% \end{figure}

% \begin{figure}[t]
%     \centering
%     \includegraphics[width=0.95\linewidth]{Images/results_time_comparison_2.pdf}
%     \caption{
%     }
%     \label{fig:patch_size_ablation_1}
% \end{figure}

\clearpage

\section{Denoising Patch Size Ablation}

\begin{figure}[h]
\centering
\includegraphics[width=0.9\linewidth]{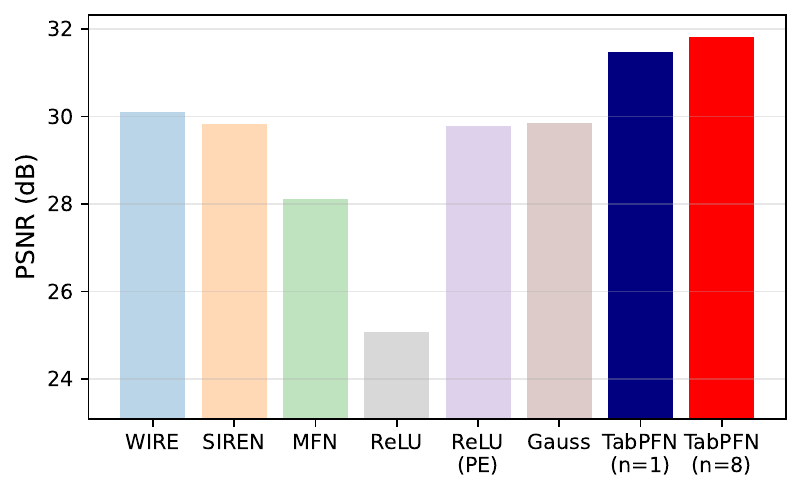}
\caption{Maximum PSNR achieved by each method across all patch sizes in \cref{fig:patch_size_ablation}. TabPFN attains the highest PSNR overall, performing strongly even with a single regressor and showing only minor additional gains with an ensemble ($n=8$).}
\label{fig:best_psnr}
\end{figure}

\begin{figure}[t]
\centering
\includegraphics[width=\linewidth]{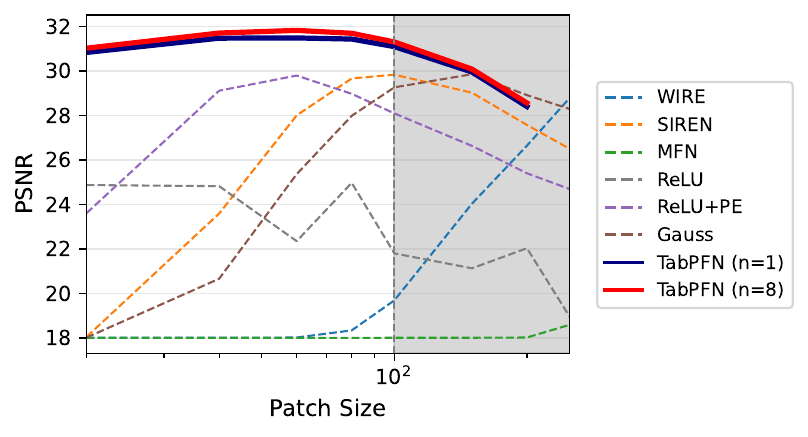}\\
\includegraphics[width=\linewidth]{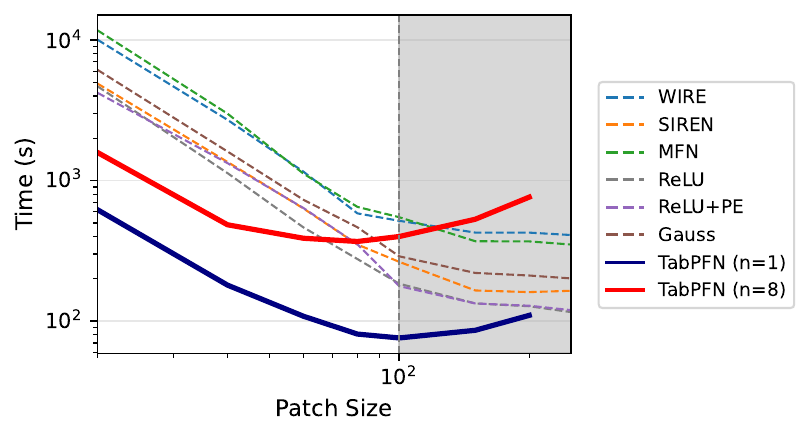}

\caption{Patch-size ablation for TabPFN-based denoising. We vary patch size and report PSNR (top) and runtime (bottom). The gray region marks patch sizes that exceed TabPFN’s 100 pixel context limit. TabPFN achieves strong PSNR even with small patches and remains stable up to its limit, while INR baselines require larger patches to reach peak performance. Runtime decreases for all methods as patch size increases, but TabPFN reaches its minimum near the recommended context limit and rises once this window is exceeded. Increasing the number of TabPFN regressors ($n=8$) yields only small PSNR gains but increases runtime.}
\label{fig:patch_size_ablation}
\end{figure}

In Section 5 we showed that TabPFN can perform image denoising competitively with leading INR methods. Because TabPFN is limited to a 10,000 sample context window, we adopted the patching strategy in Algorithm 1. Here we examine how patch size affects performance and whether TabPFN’s strong results arise from operating on small patches. We evaluate TabPFN and all baselines across a range of patch sizes. For TabPFN, we test patches from 20-200 pixels (extending beyond its recommended 100 pixel limit) and vary the number of regressors ($n=1$ and the default $n=8$) to assess the effect of ensembling. INR methods are evaluated using patch sizes from 20-1000 pixels, training each patch independently with the same hyperparameters as in Section 5. Results are shown in \cref{fig:patch_size_ablation}. TabPFN exhibits consistently strong performance within its context window and degrades only mildly once this limit is exceeded. By contrast, INR performance depends heavily on patch size, with larger patches generally required for high quality results. This sensitivity likely reflects overfitting on small patches or a need to retune hyperparameters across patch sizes for best performance - issues that TabPFN avoids. Overall, TabPFN outperforms all baselines in both PSNR and runtime. To evaluate performance independent of patch size, \cref{fig:best_psnr} reports the maximum PSNR attained by each method. Increasing the number of TabPFN regressors from $n=1$ to $n=8$ provides only marginal improvements, while substantially increasing computational cost.

\end{document}